\documentclass[11pt,a4paper]{article}
\usepackage[hyperref]{acl2018}
\usepackage{times}
\usepackage{latexsym}
\usepackage{graphicx}
\usepackage{hyperref}
\usepackage{amsmath}
\usepackage[linesnumbered,ruled]{algorithm2e}
\usepackage{multirow}

\usepackage{url}

\aclfinalcopy 

\title{Temporal Event Knowledge Acquisition via Identifying Narratives}

\author{Wenlin Yao and Ruihong Huang \\
        Department of Computer Science and Engineering\\
		Texas A\&M University\\
         {\tt \{wenlinyao, huangrh\}@tamu.edu}}

\date{}

\begin{document}
\maketitle
\begin{abstract}
Inspired by the double temporality characteristic of narrative texts, we propose a novel approach for acquiring rich temporal ``before/after'' event knowledge across sentences in 
narrative stories. The double temporality states that a narrative story often describes a sequence of events following the chronological order and therefore, the temporal order of events matches with their textual order. 
We explored narratology principles and built a weakly supervised approach that identifies 287k narrative paragraphs from three large text corpora. 
We then extracted rich temporal event knowledge from these narrative paragraphs. Such event knowledge is shown useful to improve temporal relation classification and outperform several recent neural network models on the narrative cloze task. 



\end{abstract}

\section{Introduction}

Occurrences of events, referring to changes
and actions,
show regularities. Specifically, certain events often co-occur and 
in a particular temporal order. For example, people often go to {\it work} after {\it graduation} with a degree. Such ``before/after'' temporal event knowledge can be used to recognize temporal relations between events in a document even when their local contexts do not indicate any temporal relations. Temporal event knowledge is also useful to predict an event given several other events in the context. Improving event temporal relation identification and event prediction capabilities can benefit various NLP applications, including event timeline generation, text summarization and question answering. 

While being in high demand, temporal event knowledge is lacking and difficult to obtain. Existing knowledge bases, such as Freebase~\cite{bollacker2008freebase} or Probase~\cite{wu2012probase}, often contain rich knowledge about entities, e.g., the birthplace of a person, but contain little event knowledge. Several approaches have been proposed to acquire temporal event knowledge from a text corpus, by either utilizing textual patterns \cite{chklovski2004verbocean} or building a temporal relation identifier \cite{yao2017weakly}. However, most of these approaches are limited to identifying temporal relations within one sentence. 

\begin{figure}[t]
\fbox{\begin{minipage}{19em}
\small
{\color{blue}Michael Kennedy} {\color{red}graduated} with a bachelor's degree from Harvard University in 1980. 
{\color{blue}He} {\color{red}married} {\color{blue}his} wife, Victoria, in 1981 and {\color{red}attended} law school at the University of Virginia.
After {\color{red}receiving} {\color{blue}his} law degree, {\color{blue}he} briefly {\color{red}worked} for a private law firm before joining Citizens Energy Corp.
{\color{blue}He} {\color{red}took over} management of the corporation, a non-profit firm that delivered heating fuel to the poor, from {\color{blue}his} brother Joseph in 1988.
{\color{blue}Kennedy} {\color{red}expanded} the organization goals and {\color{red}increased} fund raising.
\end{minipage}}
\fbox{\begin{minipage}{19em}
\small
{\color{blue}Beth} {\color{red}paid} the taxi driver.
{\color{blue}She} {\color{red}jumped out} of the taxi and {\color{red}headed} towards the door of her small cottage.
{\color{blue}She} {\color{red}reached into} {\color{blue}her} purse for keys.
{\color{blue}Beth} {\color{red}entered} {\color{blue}her }cottage and got {\color{red}undressed}.
{\color{blue}Beth} quickly {\color{red}showered} deciding a bath would take too long.
{\color{blue}She} {\color{red}changed} into a pair of jeans, a tee shirt, and a sweater.
Then, {\color{blue}she} {\color{red}grabbed} her bag and {\color{red}left} the cottage.
\end{minipage}}
\caption{Two narrative examples}
\label{fig:narrative_example}
\end{figure}

Inspired by the double temporality characteristic of narrative texts, we propose a novel approach for acquiring rich temporal ``before/after'' event knowledge across sentences via identifying narrative stories.  
The double temporality states that a narrative story often describes a sequence of events following the chronological order and therefore, the temporal order of events  matches with their textual order ~\cite{walsh2001fabula,riedl2010narrative,grabes2013sequentiality}.
Therefore, we can easily distill temporal event knowledge if we have identified a large collection of narrative texts. 
Consider the two narrative examples in figure \ref{fig:narrative_example}, where the top one is from a news article of New York Times and the bottom one is from a novel book. From the top one, we can easily extract one chronologically ordered event sequence \{graduated, marry, attend, receive, work, take over, expand, increase\}, with all events related to the main character Michael Kennedy. While some parts of the event sequence are specific to this story, the event sequence contains regular event temporal relations, e.g., people often \{graduate\} first and then get \{married\}, or \{take over\} a role first and then \{expand\} a goal. 
Similarly, from the bottom one, we can easily extract another event sequence \{pay, jump out, head, reach into, enter, undress, shower, change, grab, leave\} that 
contains routine actions when people take a shower and change clothes.

There has been recent research on narrative identification from blogs by building a text classifier in a supervised manner~\cite{gordon2009identifying,ceran2012hybrid}. However, narrative texts are common in other genres as well, including news articles and novel books, where little annotated data is readily available. Therefore, in order to identify narrative texts from rich sources, we develop a weakly supervised method that can quickly adapt and identify narrative texts from different genres, by heavily exploring the principles that are used to characterize narrative structures in narratology studies. It is generally agreed in narratology \cite{forster1962aspects,mani2012computational, pentland1999building,bal2009narratology} that a narrative is a discourse presenting a sequence of events arranged in their time order (the plot) and involving specific characters (the characters). First, we derive specific grammatical and entity co-reference rules to identify narrative paragraphs that each 
contains a sequence of sentences sharing the same actantial syntax structure (i.e., {\it NP VP} describing {\it a character did something}) \cite{greimas1971narrative}
and mentioning the same character. 
Then, we train a classifier using the initially identified seed narrative texts and a collection of grammatical, co-reference and linguistic features that capture the two key principles and other textual devices of narratives. Next, the classifier is applied back to identify new narratives from raw texts. The newly identified narratives will be used to augment seed narratives and the bootstrapping learning process iterates until no enough new narratives can be found. 

Then by leveraging the double temporality characteristic of narrative paragraphs, we distill general temporal event knowledge. Specifically, we extract event pairs as well as longer event sequences consisting of strongly associated events that often appear in a particular textual order in narrative paragraphs,  
by calculating 
Causal Potential \cite{beamer2009using,hu2017unsupervised} between events. 

Specifically, we obtained 19k event pairs and 25k event sequences with three to five events from the 287k narrative paragraphs we identified across three genres, news articles, novel books and blogs. Our evaluation shows that both the automatically identified narrative paragraphs and the extracted event knowledge are of high quality. 
Furthermore, the learned temporal event knowledge is shown to yield additional performance gains when used for temporal relation identification and the Narrative Cloze task. 
The acquired event temporal knowledge and the knowledge acquisition system 
are publicly available\footnote{\url{http://nlp.cs.tamu.edu/resources.html}}. 

\section{Related Work}
Several previous works have focused on acquiring temporal event knowledge from texts. VerbOcean~\cite{chklovski2004verbocean} used pre-defined lexico-syntactic patterns (e.g., ``X and then Y'') to acquire event pairs with the temporal {\it happens\_before} relation from the Web. \citet{yao2017weakly} simultaneously trained a temporal ``before/after'' relation classifier and acquired event pairs that are regularly in a temporal relation by exploring the observation that some event pairs tend to show the same temporal relation regardless of contexts.
Note that these prior works are limited to identifying temporal relations within individual sentences. In contrast, our approach 
is designed to acquire temporal relations across sentences in a narrative paragraph. Interestingly, only 195 (1\%) out of 19k event pairs acquired by our approach can be found in VerbOcean or regular event pairs learned by the previous two approaches.

Our design of the overall event knowledge acquisition also benefits from recent progress on narrative identification. \citet{gordon2009identifying} annotated a small set of paragraphs presenting stories in the ICWSM Spinn3r Blog corpus~\cite{burton2009icwsm} and trained a classifier using bag-of-words features to identify more stories. \cite{ceran2012hybrid} trained a narrative classifier using semantic triplet features on the CSC Islamic Extremist corpus. Our weakly supervised narrative identification method is closely related to 
\citet{eisenberg2017simpler}, which also explored the two key elements of narratives, the plot and the characters, in designing features with the goal of obtaining a generalizable story detector. But different from this work, our narrative identification method does not require any human annotations and can quickly adapt to new text sources.

Temporal event knowledge acquisition is related to script learning \cite{chambers2008unsupervised}, where a script consists of a sequence of events that are often temporally ordered and represent a typical scenario. 
However, most of the existing approaches on script learning \cite{chambers2009unsupervised,pichotta2016learning,granroth2016happens} were designed to identify clusters of closely related events, not to learn the temporal order between events though. 
For example, \citet{chambers2008unsupervised,chambers2009unsupervised} learned event scripts by first identifying closely related events that share an argument and then recognizing their partial temporal orders by a separate temporal relation classifier trained on the small labeled dataset TimeBank~\cite{pustejovsky2003timebank}. 
Using the same method to get training data,
\citet{jans2012skip,granroth2016happens,pichotta2016learning,wang2017integrating} applied neural networks to learn event embeddings and predict the following event in a context. Distinguished from the previous script learning works, we focus on acquiring event pairs or longer script-like event sequences with events arranged in a complete temporal order. 
In addition, recent works \cite{regneri2010learning,modi2017inscript} collected script knowledge by directly asking Amazon Mechanical Turk (AMT) to write down typical temporally ordered event sequences in a given scenario (e.g., shopping or cooking). 
Interestingly, our evaluation shows that our approach can yield temporal event knowledge that covers 48\% of human-provided script knowledge. 
\section{Key Elements of Narratives}\label{sec:narratology}
It is generally agreed in narratology \cite{forster1962aspects,mani2012computational, pentland1999building,bal2009narratology} that a narrative presents a sequence of events arranged in their time order (the plot) and involving specific characters (the characters). 


\noindent{\bf Plot}.
The plot consists of a sequence of closely related events. 
According to~\cite{bal2009narratology}, an event in a narrative often describes a ``transition from one state to another state, caused or experienced by actors''. 
Moreover, as~\citet{mani2012computational} illustrates, a narrative is often ``an account of past events in someone's life or in the development of something''. 
These prior studies suggest that sentences containing a plot event are likely to have the actantial syntax ``NP VP''\footnote{NP is Noun Phrase and VP is Verb Phrase.} \cite{greimas1971narrative} with the main verb in the past tense. 


\noindent{\bf Character}.
A narrative usually describes events caused or experienced by actors. Therefore, a narrative story often has one or two main characters, called protagonists, who are involved in multiple events and tie events together. The main character can be a person or an organization. 

\noindent{\bf Other Textual Devices}.
A narrative may contain peripheral contents other than events and characters, including time, place, the emotional and psychological states of characters etc., which do not advance the plot but provide essential information to the interpretation of the events~\cite{pentland1999building}. We use rich Linguistic Inquiry and Word Count (LIWC)~\cite{pennebaker2015development} features to capture a variety of textual devices used to describe such contents. 

\section{Phase One: Weakly Supervised Narrative Identification}
In order to acquire rich temporal event knowledge, we first develop a weakly supervised 
approach that can quickly adapt to identify narrative paragraphs from various text sources. 

\begin{figure*}[t]
 \centering
 \includegraphics[width = 4.5in]{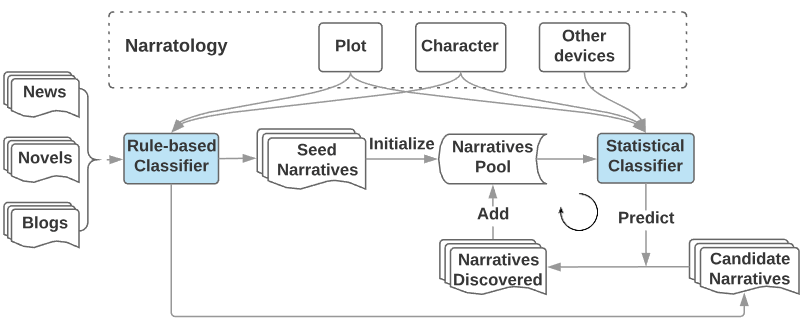}
 \caption{Overview of the Narrative Learning System}
\label{boot_pipeline}
\end{figure*}

\subsection{System Overview}
The weakly supervised method is designed to capture key elements of narratives in each of two stages. 
As shown in figure \ref{boot_pipeline}, in the first stage, we identify the initial batch of 
narrative paragraphs that satisfy strict rules and the key principles of narratives. 
Then in the second stage, we train a statistical classifier using
the initially identified seed narrative texts and a
collection of soft features for capturing the same key principles and
other textual devices of narratives. Next, the classifier
is applied to identify new narratives from raw
texts again. The newly identified narratives will be used
to augment seed narratives and the bootstrapping
learning process iterates until no enough (specifically, less than 2,000) new narratives
can be found.  
Here, in order to specialize the statistical classifier to each genre, we conduct  the learning process on news, novels and blogs separately.



\subsection{Rules for Identifying Seed Narratives}\label{sec:seed_features}

\vspace{.05in}
\noindent{\bf Grammar Rules for Identifying Plot Events}\label{sec:event_detection}.
Guided by the prior narratology studies \cite{greimas1971narrative,mani2012computational} and our observations, we use context-free grammar production rules to identify sentences that describe an event in an actantial syntax structure. Specifically, we use three sets of grammar rules to specify the overall syntactic structure of a sentence. 
First, we require a sentence to have the basic active voiced structure ``S $\rightarrow$ NP VP'' or one of the more complex sentence structures that are derived from the basic structure considering Coordinating Conjunctions (CC), Adverbial Phrase (ADVP) or Prepositional Phrase (PP) attachments\footnote{We manually identified 14 top-level sentence production rules, for example, ``S $\rightarrow$ NP ADVP VP'', ``S $\rightarrow$ PP , NP VP'' and ``S $\rightarrow$ S CC S''. Appendix shows all the rules.}. For example, in the narrative of Figure \ref{fig:narrative_example}, the sentence {\it ``Michael Kennedy earned a bachelor's degree from Harvard University in 1980.''} has the basic sentence structure ``S $\rightarrow$ NP VP'', where the ``NP'' governs the character mention of `Michael Kennedy' and the ``VP'' governs the rest of the sentence and describes a plot event. 

In addition, considering that a narrative is usually ``an account of past events in someone's life or in the development of something''~\cite{mani2012computational,dictionary2007oxford}, we require the headword of the VP to be 
in the past tense. Furthermore, the subject of the sentence is meant to represent a character. Therefore, we specify 12 grammar rules\footnote{The example NP rules include ``NP $\rightarrow$ NNP'', ``NP $\rightarrow$ NP CC NP'' and ``NP $\rightarrow$ DT NNP''.} to require the sentence subject noun phrase to have a simple structure and have a proper noun or pronoun as its head word.  

For seed narratives, we consider paragraphs containing at least four sentences and we require 60\% or more sentences to satisfy the sentence structure specified above. We also require a narrative paragraph to contain no more than 20\% of sentences that are interrogative, exclamatory or dialogue, which normally do not contain any plot events.
The specific parameter settings are mainly determined based on our observations and analysis of narrative samples. The threshold of 60\% for ``sentences with actantial structure'' was set to reflect the observation that sentences in a narrative paragraph usually (over half) have an actantial structure. A small portion (20\%) of interrogative, exclamatory or dialogue sentences is allowed to reflect the observation that many paragraphs are overall narratives even though they may contain 1 or 2 such sentences, so that we achieve a good coverage in narrative identification.

\vspace{.05in}
\noindent{\bf The Character Rule}.
A narrative usually has a protagonist character that appears in multiple sentences and ties a sequence of events, therefore, we also specify a rule requiring a narrative paragraph to have a protagonist character. 
Concretely, inspired by \citet{eisenberg2017simpler}, we applied the named entity recognizer \cite{finkel2005incorporating} and entity coreference resolver \cite{lee2013deterministic} from the CoreNLP toolkit \cite{manning2014stanford} to identify the longest entity chain in a paragraph that has at least one mention recognized as a {\it Person} or {\it Organization}, or a gendered pronoun. Then we calculate the normalized length of this entity chain by dividing the number of entity mentions by the number of sentences in the paragraph. We require the normalized length of this longest entity chain to be $\ge 0.4$, meaning that 40\% or more sentences in a narrative mention a character
\footnote{40\% was chosen to reflect that a narrative paragraph often contains a main character that is commonly mentioned across sentences (half or a bit less than half of all the sentences).}.   


\subsection{The Statistical Classifier for Identifying New Narratives}
Using the seed narrative paragraphs identified in the first stage as positive instances, we train a statistical classifier to continue to identify more narrative paragraphs that may not satisfy the specific rules.
We also prepare negative instances to compete with positive narrative paragraphs in training. Negative instances are paragraphs that are not likely to be narratives and do not present a plot or protagonist character, but are similar to seed narratives in others aspects. Specifically, similar to seed narratives, we require a non-narrative paragraph to contain at least four sentences with 
no more than 20\% of sentences being interrogative, exclamatory or dialogue; but in contrast to seed narratives, a non-narrative paragraph should contain 30\% of or fewer sentences that have the actantial sentence structure, where the longest character entity chain should not span over 20\% of sentences. 
We randomly sample such non-narrative paragraphs that are five times of narrative paragraphs
\footnote{We used the skewed pos:neg ratio of 1:5 in all bootstrapping iterations to reflect the observation that there are generally many more non-narrative paragraphs than narrative paragraphs in a document.}. 

In addition, since it is infeasible to apply the trained classifier to all the paragraphs
in a large text corpus, such as the Gigaword corpus \cite{graff2003english}, 
we identify candidate narrative paragraphs and only apply the statistical classifier to these candidate paragraphs. Specifically, we require a candidate paragraph to satisfy all the constraints used for identifying seed narrative paragraphs but contain only 30\%\footnote{This value is half of the corresponding thresshold used for identifying seed narrative paragraphs.} 
or more sentences with an actantial structure and have the longest character entity chain spanning over 20\%\footnote{This value is half of the corresponding thresshold used for identifying seed narrative paragraphs.} 
of or more sentences.   

We choose Maximum Entropy~\cite{berger1996maximum} as the classifier. 
Specifically, we use the MaxEnt model implementation in the LIBLINEAR library\footnote{\url{https://www.csie.ntu.edu.tw/\~cjlin/liblinear/}}~\cite{fan2008liblinear} with default parameter settings. Next, we describe the features used to capture the key elements of narratives. 


\vspace{.05in}
\noindent{\bf Features for Identifying Plot Events:} 
Realizing that grammar production rules are effective in identifying sentences that contain a plot event, we encode all the production rules as features in the statistical classifier. 
Specifically, for each narrative paragraph, we use the frequency of all syntactic production rules as features. 
Note that the bottom level syntactic production rules 
have the form of POS tag $\rightarrow$ WORD and contain a lexical word, which made these rules dependent on specific contexts of a paragraph. Therefore, we exclude these bottom level production rules from the feature set 
in order to model generalizable narrative elements rather than specific contents of a paragraph. 

In addition, to capture potential event sequence overlaps between new narratives and the already learned narratives, we build a verb bigram language model using verb sequences extracted from the learned narrative paragraphs and calculate the perplexity score (as a feature) of the verb sequence in a candidate narrative paragraph.
Specifically, we calculate the perplexity score of an event sequence that is normalized by the number of events,  
$PP(e_1,...,e_N) = \sqrt[N]{\prod_{i=1}^{N}\frac{1}{P(e_i|e_{i-1})}}$, 
where $N$ is the total number of events in a sequence and $e_i$ is a event word. We approximate $P(e_i|e_{i-1}) = \frac{C(e_{i-1}, e_{i})}{C(e_{i-1})}$, where $C(e_{i-1})$ is the number of occurrences of $e_{i-1}$ and $C(e_{i-1}, e_{i})$ is the number of co-occurrences of $e_{i-1}$ and $e_{i}$.
$C(e_{i-1}, e_{i})$ and $C(e_{i-1})$ are calculated based on all event sequences from known narrative paragraphs. 

\vspace{.05in}
\noindent{\bf Features for the Protagonist Characters:} 
We consider the longest three coreferent entity chains in a paragraph 
that have at least one mention recognized as a {\it Person} or {\it Organization}, or a gendered pronoun. Similar to the seed narrative identification stage, we obtain the normalized length of each entity chain by dividing the number of entity mentions with the number of sentences in the paragraph. In addition, we also observe that a protagonist character appears frequently in the surrounding paragraphs as well, therefore, we calculate the normalized length of each entity chain based on its presences in the target paragraph as well as one preceding paragraph and one following paragraph. We use 6 normalized lengths (3 from the target paragraph
\footnote{Specifically, the lengths of the longest, second longest and third longest entity chains.}
and 3 from surrounding paragraphs) as features. 



\vspace{.05in}
\noindent{\bf Other Writing Style Features:} We create a feature for each semantic category in the Linguistic Inquiry and Word Count (LIWC) dictionary~\cite{pennebaker2015development}, and the feature value is the
total number of occurrences of 
all words in that category. These LIWC features capture presences of certain types of words, such as words denoting relativity (e.g., motion, time, space) and words referring to psychological processes (e.g., emotion and cognitive). 
In addition, we encode Parts-of-Speech (POS) tag frequencies as features as well which have been shown effective in identifying text genres and writing styles.

\begin{table}[t]
\small
\begin{center}
\begin{tabular}{ |l|cccccc|}\hline

& 0 (Seeds) & 1  & 2 & 3 & 4 & Total\\ \hline
News &  20k & 40k & 12k & 5k & 1k & 78k \\ 
Novels &  75k & 82k & 24k & 6k & 2k & 189k \\
Blogs &  6k & 10k & 3k & 1k & - & 20k \\\hline
Sum & 101k & 132k & 39k & 12k & 3k & 287k\\\hline
\end{tabular}
\end{center}
\caption{Number of new narratives generated after each bootstrapping iteration}
\label{bootstrapping}
\end{table}

\subsection{Identifying Narrative Paragraphs from Three Text Corpora}

Our weakly supervised system is based on the principles shared across all narratives, so it can be applied to different text sources for identifying narratives. 
We considered three types of texts: 
(1) {\bf News Articles}. News articles contain narrative paragraphs 
to describe the background of an important figure or to provide details for a significant event.
We use English Gigaword 5th edition~\cite{graff2003english,napoles2012annotated}, which  contains 10 million news articles.
(2) {\bf Novel Books}.
Novels contain rich narratives to describe actions by characters. BookCorpus~\cite{zhu2015aligning}
is a large collection of free novel books written by unpublished authors, which contains 11,038 books of 16 different sub-genres (e.g., Romance, Historical, Adventure, etc.). 
(3) {\bf Blogs}.
Vast publicly accessible blogs also contain narratives 
because ``personal life and experiences'' is a primary topic of blog posts~\cite{lenhart2006bloggers}. 
We use the Blog Authorship Corpus~\cite{schler2006effects} collected from the blogger.com website, which consists of 680k posts written by thousands of authors. 
We applied the Stanford CoreNLP tools~\cite{manning2014stanford} 
 to the three text corpora 
to obtain POS tags, parse trees, named entities, coreference chains, etc. 

In order to combat semantic drifts \cite{mcintosh2009reducing} in bootstrapping learning, we set the initial selection confidence score produced by the statistical classifier at 0.5 and increase it by 0.05 after each iteration. The bootstrapping system runs for four iterations and learns 287k narrative paragraphs in total. 
Table \ref{bootstrapping} shows the number of narratives that were obtained in the seeding stage and in each bootstrapping iteration from each text corpus. 

\section{Phase Two: Extract Event Temporal Knowledge from Narratives}\label{sec:CausalEventPairs}
Narratives we obtained from the first phase may describe specific stories and contain uncommon events or event transitions. Therefore, we apply Pointwise Mutual Information (PMI) based statistical metrics to measure strengths of event temporal relations in order to identify general knowledge that is not specific to any particular story. Our goal is to learn event pairs and longer event chains with events completely ordered in the temporal ``before/after'' relation.  

First, by leveraging the double temporality characteristic of narratives, we only consider event pairs and longer event chains with 3-5 events that have occurred as a segment in at least one event sequence extracted from a narrative paragraph. Specifically, we extract the event sequence (the plot) from
a narrative paragraph by finding the main event in each
sentence and chaining the main events\footnote{We only consider main events that are in base verb forms or in the past tense, by requiring their POS tags to be VB, VBP, VBZ or VBD.} according to their textual order. 

Then we rank candidate event pairs based on two factors, how strongly associated two events are and how common they appear in a particular temporal order. 
We adopt the existing metric, 
Causal Potential (CP), which has been applied to acquire causally related events \cite{beamer2009using} and exactly measures the two aspects. Specifically, the CP score of an event pair is calculated using the following equation:
\begin{equation}
\label{CP_score}
\small
cp(e_i, e_j) = pmi(e_i,e_j) + log\frac{P(e_i \rightarrow e_j)}{P(e_j \rightarrow e_i)}
\end{equation} 
where, the first part refers to the Pointwise Mutual Information (PMI) between two events and the second part measures the relative ordering or two events. $P(e_i \rightarrow e_j)$ refers to the probability that $e_i$ occurs before $e_j$ in a text, which is proportional to the raw frequency of the pair. 
PMI measures the 
association strength of two events, formally, $pmi(e_i,e_j) = log\frac{P(e_i,e_j)}{P(e_i)P(e_j)}$, 
$P(e_i) = \frac{C(e_i)}{\sum_{x}C(e_x)}$ and $P(e_i,e_j) = \frac{C(e_i,e_j)}{\sum_{x}\sum_{y}C(e_x, e_y)}$, 
where, $x$ and $y$ refer to all the events in a corpus,  $C(e_i)$ is  the number of occurrences 
of $e_i$, 
$C(e_i, e_j)$ is the number 
of co-occurrences of $e_i$ and $e_j$. 

While each candidate pair of events should have appeared consecutively as a segment in at least one narrative paragraph, when calculating the CP score, we consider event co-occurrences even when two events are not consecutive in a narrative paragraph but have one or two other events in between. Specifically, the same as in \cite{hu2017inferring}, we calculate separate CP scores based on event co-occurrences with zero (consecutive), one or two events in between, and use the weighted average CP score for ranking an event pair, formally, $CP(e_i, e_j) = \sum_{d=1}^{3}\frac{cp_d(e_i, e_j)}{d}$.

Then we rank longer event sequences based on CP scores for individual event pairs that are included in an event sequence. However, an event sequence of length $n$ is more than $n-1$ event pairs with any two consecutive events as a pair. 
We prefer event sequences that are coherent overall, where the events that are one or two events away are highly related as well.  Therefore, 
we define the following metric to measure the quality of an event sequence:
\begin{equation}
\small
CP(e_1,e_2,\cdots,e_n) = \frac{\sum_{d=1}^{3}\sum_{j=1}^{n-d}\frac{CP(e_j, e_{j+d})}{d}}{n-1}.
\end{equation}

\section{Evaluation}

\subsection{Precision of Narrative Paragraphs}
From all the learned narrative paragraphs, we randomly selected 150 texts, with 25 texts selected from narratives learned in each of the two stages (i.e., seed narratives and bootstrapped narratives) using each of the three text corpora (i.e., news, novels, and blogs). 
Following the same definition ``A story is a narrative of events arranged in their time sequence''~\cite{forster1962aspects,gordon2009identifying}, two human adjudicators were asked to judge whether each text is a narrative or a non-narrative. 
In order to obtain high inter-agreements, before the official annotations, we trained the two annotators 
for several iterations. Note that the texts we used in training annotators are different from the final texts we used for evaluation purposes. The overall kappa inter-agreement between the two annotators is 0.77.

Table \ref{tab:narrative_acc} shows the precision of narratives learned in the two stages using the three corpora. We determined
that a text is a correct narrative if both annotators labeled it as a narrative. 
We can see that on average, the rule-based classifier achieves the precision of 88\% on initializing seed narratives and the statistical classifier achieves the precision of 84\% on bootstrapping new ones. Using narratology based features enables the statistical classifier to extensively learn new narrative, and meanwhile maintain a high precision. 

\begin{table}[t]
\centering
\begin{tabular}{l|rr}
\hline
Narratives & Seed & Bootstrapped\\\hline
News & 0.84 & 0.72\\
Novel & 0.88 & 0.92\\
Blogs & 0.92 & 0.88 \\ \hline
AVG & 0.88 & 0.84 \\
\hline
\end{tabular}
\caption{\label{tab:narrative_acc} Precision of narratives based on human annotation}
\end{table}

\begin{table}[t]
\small
\centering
\begin{tabular}[center]{|c|l|}\hline
\multirow{3}{*}{pairs}  & graduate $\rightarrow$ teach (5.7), meet $\rightarrow$ marry (5.3) \\
                        & pick up $\rightarrow$ carry (6.3), park $\rightarrow$ get out (7.3) \\
                        & turn around $\rightarrow$ face (6.5), dial $\rightarrow$ ring (6.3) \\
\hline
\multirow{6}{*}{chains} & drive $\rightarrow$ park $\rightarrow$ get out  (7.8) \\
                        & toss $\rightarrow$ fly $\rightarrow$ land  (5.9) \\
                        & grow up $\rightarrow$ attend $\rightarrow$ graduate $\rightarrow$ marry (6.9) \\
                        & contact $\rightarrow$ call $\rightarrow$ invite $\rightarrow$ accept (4.2) \\
                        & knock $\rightarrow$ open $\rightarrow$ reach $\rightarrow$ pull out $\rightarrow$ hold (6.0)\\
\hline
\end{tabular}
\caption{Examples of 
event pairs and chains (with CP scores). 
$\rightarrow$ represents {\it before} relation.}
\label{tab:pair_chain_examples}
\end{table}

\subsection{Precision of Event Pairs and Chains}
To evaluate the quality of the extracted event pairs and chains, we randomly sampled 20 pairs (2\%) from every 1,000 event pairs up to the top 18,929 pairs with CP score $\geq$ 2.0 (380 pairs selected in total), and 10 chains (1\%) from every 1,000 up to the top 25,000 event chains\footnote{It turns out that many event chains have a high CP score close to 5.0, so we decided not to use a cut-off CP score of event chains but simply chose to evaluate the top 25,000 event chains.} 
(250 chains selected in total). The average CP scores for all event pairs and all event chains we considered are 2.9 and 5.1 respectively. Two human adjudicators were asked to judge whether or not events are likely to occur in the temporal order shown. For event chains, we have one additional criterion requiring that events form a coherent sequence overall. An event pair/chain is deemed correct if both annotators labeled it as correct. 
The two annotators achieved kappa inter-agreement scores of 0.71 and 0.66, on annotating event pairs and event chains respectively.

\begin{figure}[t]
 \centering
 \includegraphics[width = 2.6in]{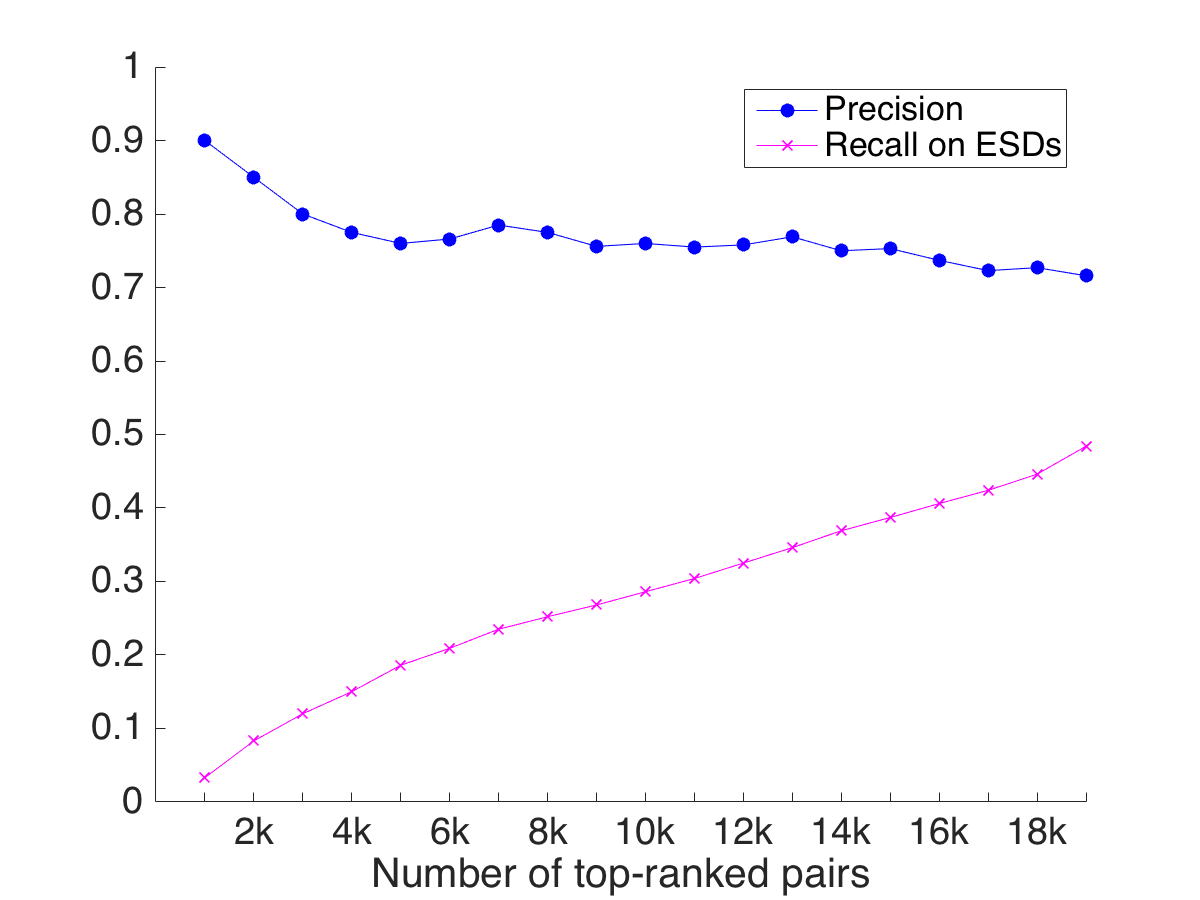}
 \caption{Top-ranked event pairs evaluation}
\label{pair_eval}
\end{figure}

\begin{table}[t]
\small
\begin{center}
\begin{tabular}{ |l|ccccc|}\hline

\# of top chains & 5k & 10k & 15k & 20k & 25k \\ \hline
Precision & 0.76 & 0.8 & 0.75 & 0.73 & 0.69 \\\hline
\end{tabular}
\end{center}
\caption{Precision of top-ranked event chains}
\label{tab:chain_eval}
\end{table}

As we know, coverage on acquired knowledge is often hard to evaluate because we do not have a complete  knowledge base to compare to. Thus, we propose a pseudo recall metric to evaluate the coverage of event knowledge we acquired. \citet{regneri2010learning} collected Event Sequence Descriptions (ESDs) of several types of human activities (e.g., baking a cake, going to the theater, etc.) using crowdsourcing. Our first pseudo recall score is calculated based on how many consecutive event pairs in human-written scripts can be found in our top-ranked event pairs. 
Figure \ref{pair_eval} illustrates the precision of top-ranked pairs based on human annotation and the pseudo recall score based on ESDs. We can see that about 75\% of the top 19k event pairs are correct, which captures 48\% of human-written script knowledge in ESDs. 
In addition, table \ref{tab:chain_eval} shows the precision of top-ranked event chains with 3 to 5 events. Among the top 25k event chains, about 70\% are correctly  ordered with the temporal ``after'' relation.  Table \ref{tab:pair_chain_examples} shows several examples of event pairs and chains. 

\subsection{Improving Temporal Relation Classification by Incorporating Event Knowledge}
To find out whether the learned temporal event knowledge can help with improving temporal relation classification performance, 
we conducted experiments on a benchmark dataset - TimeBank corpus v1.2, which contains 2308 event pairs that are annotated with 14 temporal relations \footnote{Specifically, the 14 relations are {\it simultaneous, before, after, ibefore, iafter, begins, begun by, ends, ended by, includes, is included, during, during inv, identity}}. 

To facilitate direct comparisons, we 
used the same state-of-the-art temporal relation classification system as described in our previous work \citet{prafulla2017sequential} and considered all the 14 relations in classification. \citet{prafulla2017sequential} forms three sequences (i.e., word forms, POS tags, and dependency relations) of context words that align with the dependency path between two event mentions and uses three bi-directional LSTMs to get the embedding of each sequence. The final fully connected layer maps the concatenated embeddings of all sequences to 14 fine-grained temporal relations.  We applied the same model here, but if an event pair appears in our learned list of event pairs, we concatenated the CP score of the event pair 
as additional evidence in the final layer. To be consistent with \citet{prafulla2017sequential}, we used the same train/test splitting, the same parameters for the neural network and only considered intra-sentence event pairs. Table \ref{tab:timebankperf} shows that by incorporating our learned event knowledge,  
the overall prediction accuracy was improved by 1.1\%. Not surprisingly, out of the 14 temporal relations, the performance on the relation {\it before} was improved the most by 4.9\%.


\begin{table}[t]
\centering
\begin{tabular}{l|r}
\hline
Models & Acc.(\%)\\
\hline
\citet{prafulla2017sequential} & 51.2\\
+ CP score & {\bf 52.3}\\
\hline
\end{tabular}
\caption{\label{tab:timebankperf} Results on TimeBank corpus}
\end{table}

\begin{table}[t]
\centering
\begin{tabular}{l|r}
\hline
Method & Acc.(\%)\\
\hline
\cite{chambers2008unsupervised} & 30.92\\
\cite{granroth2016happens} & 43.28\\
\cite{pichotta2016learning} & 43.17\\
\cite{wang2017integrating} & 46.67\\\hline
Our Results & {\bf 48.83}\\
\hline
\end{tabular}
\caption{\label{tab:performances} Results on MCNC task}
\end{table}
\subsection{Narrative Cloze}
Multiple Choice version of the Narrative Cloze task (MCNC) proposed by~\citet{granroth2016happens,wang2017integrating}, aims to evaluate understanding of a script by predicting the next event given several context events. 
Presenting a chain of contextual events $e_1, e_2, ..., e_{n-1}$, the task is to select the next event from 
five event candidates, one of which is correct 
and the others are randomly sampled elsewhere in the corpus. 
Following the same settings of \citet{wang2017integrating} and \citet{granroth2016happens}, we 
adapted the dataset (test set) of 
\citet{chambers2008unsupervised} 
to the multiple choice setting. 
The dataset contains 69 documents and 349 multiple choice questions. 

We calculated a PMI score between a candidate event and each context event $e_1, e_2, ..., e_{n-1}$ based on event sequences extracted from our learned 287k narratives and we chose the event that have the highest sum score of all individual PMI scores. 
Since the prediction accuracy on 349 multiple choice questions depends on the random initialization of four negative candidate events, we ran the experiment 10 times and took the average accuracy as the final performance.

Table \ref{tab:performances} shows the comparisons of our results with the performance of several previous models, which were all trained with 1,500k event chains extracted from the NYT portion of the Gigaword corpus~\cite{graff2003english}. Each event chain consists of a sequence of verbs sharing an actor within a news article. 
Except~\citet{chambers2008unsupervised}, other recent models 
utilized more and more sophisticated neural language models. 
\citet{granroth2016happens} proposed a two layer neural network model that learns  embeddings of event predicates and their arguments for predicting the next event.
\citet{pichotta2016learning} introduced a LSTM-based language model for event prediction. 
\citet{wang2017integrating} used dynamic memory as attention in LSTM for prediction. 
It is encouraging that by using event knowledge extracted from automatically identified narratives, we achieved the best event prediction performance, which is 2.2\% higher than the best neural network model. 

\section{Conclusions}
This paper presents a novel approach for leveraging the double temporality characteristic of narrative texts and acquiring temporal event knowledge across sentences in narrative paragraphs. 
We developed a weakly supervised system that explores narratology principles and identifies narrative texts from three text corpora of distinct genres. 
The temporal  
event knowledge distilled from narrative texts 
were shown useful to improve temporal relation classification and outperform several neural language models on the narrative cloze task.  
For the future work, we plan to expand event temporal knowledge acquisition by dealing with event sense disambiguation and event synonym identification (e.g., drag, pull and haul).  


\section{Acknowledgments}
We thank our anonymous reviewers for
providing insightful review comments.

\bibliography{acl2018}
\bibliographystyle{acl_natbib}

\appendix

\section{Appendix}
\label{sec:appendix}
Here is the full list of grammar rules for identifying plot events in the seeding stage (Section \ref{sec:event_detection}). 

\noindent{Sentence rules (14)}:


S $\rightarrow$ S CC S

S $\rightarrow$ S PRN CC S

S $\rightarrow$ NP VP

S $\rightarrow$ NP ADVP VP

S $\rightarrow$ NP VP ADVP

S $\rightarrow$ CC NP VP


S $\rightarrow$ PP NP VP


S $\rightarrow$ NP PP VP


S $\rightarrow$ PP NP ADVP VP


S $\rightarrow$ ADVP S NP VP


S $\rightarrow$ ADVP NP VP


S $\rightarrow$ SBAR NP VP


S $\rightarrow$ SBAR ADVP NP VP


S $\rightarrow$ CC ADVP NP VP

\noindent{Noun Phrase rules (12)}:

NP $\rightarrow$ PRP

NP $\rightarrow$ NNP

NP $\rightarrow$ NNS

NP $\rightarrow$ NNP NNP

NP $\rightarrow$ NNP CC NNP

NP $\rightarrow$ NP CC NP

NP $\rightarrow$ DT NN

NP $\rightarrow$ DT NNS

NP $\rightarrow$ DT NNP

NP $\rightarrow$ DT NNPS

NP $\rightarrow$ NP NNP

NP $\rightarrow$ NP NNP NNP

\end{document}